\documentclass[10pt,twocolumn]{article} 
\usepackage{simpleConference}
\usepackage{times}
\usepackage{graphicx}

\usepackage{amsmath,amssymb} 
\usepackage{color}
\usepackage{siunitx}
\usepackage[table]{xcolor}
\usepackage{caption}
\usepackage{epsfig} 
\usepackage{array}
\usepackage{booktabs}
\usepackage{multirow}
\newcommand{\head}[1]{\textnormal{\textbf{#1}}}

\begin{document}

\title{A Deeper Insight into the UnDEMoN: Unsupervised Deep Network for Depth and Ego-Motion Estimation}

\author{Madhu Babu V, Anima Majumder, Kaushik Das and  Swagat Kumar \\
\\
Technical Report \\
TATA Consultancy Services, Bangalore, India. \\
\today
\\
\\
 (madhu.vankadari, anima.majumder, kaushik.da, swagat.kumar)@tcs.com  \\
}

\maketitle
\thispagestyle{empty}

\begin{abstract}
This paper presents an unsupervised deep learning framework called UnDEMoN  for estimating dense depth map and 6-DoF camera pose information directly from monocular images. The proposed network is trained using unlabeled monocular stereo image pairs and is shown to provide superior performance in depth and ego-motion estimation compared to the existing state-of-the-art. These improvements are achieved by introducing a new objective function that aims to minimize spatial as well as temporal reconstruction losses simultaneously. These losses are defined using bi-linear sampling kernel and penalized using the Charbonnier penalty function.  The objective function, thus created, provides robustness to image gradient noises thereby improving the overall estimation accuracy without resorting to any coarse to fine strategies which are currently prevalent in the literature. Another novelty lies in the fact that we combine a disparity-based depth estimation network with a pose estimation network to obtain absolute scale-aware 6 DOF Camera pose and superior depth map. The effectiveness of the proposed approach is demonstrated through performance comparison with the existing supervised and unsupervised methods on the KITTI driving dataset. 
\end{abstract}
\section{Introduction}

Estimating depth and pose information from images is a challenging problem which finds use in several applications such as, autonomous navigation \cite{handa2014benchmark}, 3D scene reconstruction \cite{stereoscan}, augmented and virtual reality \cite{pose_ar} \cite{taylor2016efficient}. The recent success of deep learning methods has prompted many researchers to apply these techniques to solve this problem. For instance, there are several deep networks that estimate depth \cite{eigen} \cite{demon} \cite{cnn_slam} as well as pose \cite{deepvo} \cite{kendall2015posenet} \cite{vinet} directly from images using supervised learning framework that necessitates availability of explicit ground truth information which is usually difficult to collect in many real world applications. This limitation can be overcome by adopting an unsupervised learning framework as in \cite{mono_depth} \cite{sfm_learner} \cite{garg} where the depth and pose estimation problem is solved by posing it as an image reconstruction problem instead of directly regressing them to their respective ground truths. In \cite{mono_depth}, the authors use consistency between left-to-right and right-to-left disparities to obtain dense depth map from monocular images. This work relied only on spatial reconstruction losses for depth estimation which might not work in many cases, for instance, a case where the occlusions are present in one of the camera views making it impossible to reconstruct those occluded pixels.  We believe that such cases can be dealt with by going forward or backward in time sequence of captured images. Such temporal consistency was first used in \cite{sfm_learner} where the authors train a deep network with a sequence of monocular images (\emph{snippet} of $n$-images) for depth and ego-motion estimation. Since this network is trained only on monocular images, it lacks absolute scale information and hence, requires a post processing stage to retrieve it, thereby limiting its practical application in the real world. This limitation is overcome by  Li et al in  \cite{undeepvo} where they use a depth prediction network and combined it with a pose estimation network to obtain scale-aware depth and pose. However, estimating depth directly has its own limitations which can be remedied by using a disparity-based depth estimation network. The advantages of using a disparity-based depth estimation are as follows. First, disparity allows representation of points at infinite horizon and accounts for growing localization uncertainty with increasing distance. Second, the disparity is more sensitive to camera motion which makes it easier to account for even small changes in the surrounding and the objects closer to the camera. Finally, unlike depth, disparity is independent of camera calibration parameters making it more useful for real-world applications.

\begin{figure}[!t]
\centering
\includegraphics[scale=0.24]{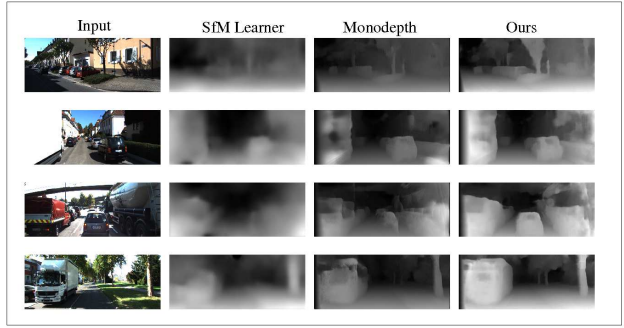}
\caption{\scriptsize The UnDEMoN results on KITTI eigen split \cite{eigen} dataset compared to the state-of-the-art methods like MonoDepth \cite{mono_depth} and SfMLearner \cite{sfm_learner}. As one can observe, our proposed method provides better depth estimate compared to these methods. We are able to detect far objects and large objects more clearly even without any post processing step used in \cite{mono_depth} \cite{sfm_learner}. }
\label{fig:depth_results}
\end{figure}

\begin{figure*}
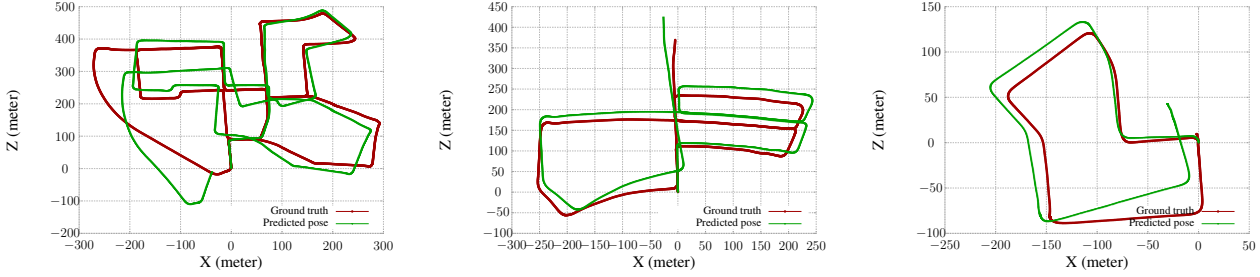

\centering
\begin{tabular}{ccc}
\scalebox{0.420}{\input{./pose_data1.tex}}&
\scalebox{0.420}{\input{./pose_data5.tex}}&
\scalebox{0.420}{\input{./pose_data7.tex}}\\
\\
\end{tabular}
\caption{The UnDEMoN estimated camera pose trajectory obtained from sequences 00, 05 and 07 respectively compared with the ground truth.}
\label{fig:pose_results}
\end{figure*}

In this paper, we extend the work presented in \cite{undeepvo} by introducing a novel objective function based on Charbonnier penalty to combine spatial and temporal reconstruction losses. The Charbonnier penalty is a differentiable variant of the absolute value function which has been shown to be robust against outliers and gradient noises \cite{charbonnier_loss}. While it has been extensively used in optical flow estimation \cite{ren2017unsupervised} \cite{charbonnier_loss}, its application to depth and pose estimation is not reported so far, thereby making it a novel contribution made in this paper. This property of Charbonnier function helps us in achieving higher accuracy without using any explicit coarse-to-fine strategies which are commonly used in literature to deal with image gradient noises \cite{garg}. Secondly, we use disparity-based depth estimation network to overcome some of the limitations of direct depth estimation carried out in \cite{undeepvo}. Third, the temporal reconstruction loss is computed using a $n$-image snippet instead of 2-image snippet, by taking the cue from Zhou et al. \cite{sfm_learner}, to make the depth bi-directionally consistent in the temporal domain. These modifications lead to significant improvement over the state-of-the-art in both depth and pose estimation as shown in Figure \ref{fig:depth_results}. As one can observe, we are able to detect farther objects and larger objects in the scene more clearly compared to the state-of-the-art methods like MonoDepth \cite{mono_depth} and SfMLearner \cite{sfm_learner}. We also show later that our depth estimation results are better than those reported for UnDeepVO \cite{undeepvo} which is the most recent work in this category. In addition, the pose estimation performance is found to be superior to other monocular methods such as SfMLearner\cite{sfm_learner} and VISO\_M \cite{libviso} and comparable with stereo based VO methods such as VISO\_S \cite{libviso}.  The estimated pose trajectories obtained with the proposed network are shown in Figure \ref{fig:pose_results} along with their respective ground truths. This new deep network framework is named as ``Unsupervised Depth and Ego-Motion Network" or simply, ``UnDEMoN" indicating an improvement over the ``DEMoN" \cite{demon} framework that uses a supervised learning approach. In short, UnDEMoN is an end-to-end unsupervised deep learning framework for estimating depth and ego-motion from monocular images. The source code, trained models and results will be made available on-line post the acceptance of this paper. 

The rest of this paper is organized as follows. A brief overview of related work is provided in the next section. The proposed architecture and the detailed method for training and testing the network is described in Section \ref{sec:meth}. The performance analysis and results are provided in Section \ref{sec:expt}. An extensive analysis on the choice of parameters and cost function is performed as ablation studies in Section \ref{sec:ablation}. Finally, Section \ref{sec:conc} concludes the paper.

\section{Related Work} \label{sec:relw}

During last few years, deep learning techniques have created new benchmarks in almost all areas of computer vision, including depth and pose estimation using monocular and stereo images. One of the earliest work in this field was carried out by Mayer et al. \cite{disp_net} who used a Convolutional deep network, called DispNet to directly predicts disparity  using stereo image pairs using a supervised learning method that relied on the availability of a large training dataset with ground truths. Similarly, Kendall et al.\cite{kendall2015posenet} use supervised training for a Convolutional neural network to predict 6-DoF camera pose directly from monocular images. This work was further extended by Li et al. \cite{li2017indoor} who used raw RGBD images with explicit depth information to improve the pose estimation accuracy. In another work called DeepVO by Wang et al. \cite{deepvo}, the authors used a recurrent version of CNN  that uses LSTM to estimate pose in the context of visual odometry.  The limitation of all these supervised methods lie in the fact that such large datasets with ground truth information are difficult to obtain in many practical applications. This limitation can be overcome by using unsupervised learning approaches as is demonstrated by later researchers. For instance, Garg et al. \cite{garg} solve the depth estimation problem in an unsupervised manner by posing it as an image reconstruction error using the opposite stereo images. This work was further improved by Godard et al. \cite{mono_depth} who used left-right consistency to make disparity bi-directionally consistent in the spatial domain. The limitation of having opposite stereo images for training is remedied by Zhou et al. \cite{sfm_learner} who solve the same problem by considering only temporal reconstruction error that is computed using $n$ temporally aligned snippets of monocular images. The temporal reconstruction losses are obtained by combining a depth estimation network with a pose estimation network. As it is well known, the use of monocular images will lead to the loss of absolute scale information in their predictions. This is solved by Li et al. \cite{undeepvo} who combine both spatial and temporal reconstruction losses to directly predict scale-aware depth and pose directly from monocular stereo images. We extend this work further by proposing several modifications including a novel objective function that significantly improves the accuracy of depth and pose estimation as will be demonstrated later in this paper. 

\section{UnDEMoN for Depth Estimation and Camera Pose Prediction}
\label{sec:meth}
In this section, we describe the proposed network architecture and the unsupervised learning framework for estimating depth and camera pose from stereo monocular images. We also define several notations and symbols which will be used throughout the paper. We first describe the generic method for depth and pose estimation in the first three subsections followed by the network architecture and various objective functions used for training the network. 

\subsection{Depth estimation from monocular image} \label{depth_estimation_sec}

\begin{figure}[h]
\centering
\includegraphics[height=4.5cm]{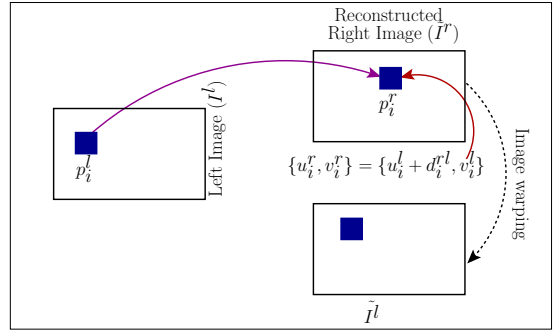} 
\caption{\scriptsize Illustrates the technique of differentiable image warping. 
Given a image point $p_i^l$ at left image $I^l$, we first project it onto the spatial-reconstructed right image $\tilde{I^r}$ using right-left disparities $d^{rl}$. Then we use bilinear interpolation for getting the warped image $\tilde{I}^l$.} 
\label{fig:warp}
\end{figure}

One of the main objectives of our approach is to learn a function which can predict per-pixel depth map $\tilde d$ in an unsupervised manner using monocular images denoted by the symbol $I$. Instead of learning a regression model using the ground truth depth, we try to predict the dense correspondence field ($d$) between left and right images. Assuming the images are stereo rectified, our network estimates the disparity $d$ expressed as a scalar per pixel value. This disparity can be used for reconstructing one image given the other in the stereo pair. As it is shown in the Figure \ref{fig:warp},
let us assume that a pixel location in the left image ${I^l}$ is represented by the notation  $p_i^l = \{u_i^l,v_i^l\}, i = 1,2,...,N$ where $N$ is the total number of pixels in the image. Let the disparity from the right to the left image be denoted by  $D^{rl} = \{d_i^{rl}, i = 1,2,..., N$\}. 

Similarly, let us assume that the right image reconstructed from the above left image and the left-to-right disparity be denoted by $\tilde{I}^{r}$ and the corresponding pixel location for this reconstructed image could be expressed as $p_i^r = \{u_i^l+d_i^{rl}, v_i^l\}$. Similarly, the left image could be reconstructed (denoted as $\tilde{I}^l$ from the right image $I^r$ and disparity $D^{lr}$. The disparity values for each pixel can be used for computing the absolute depth for that pixel using the formula: $\textbf{d}_i = \frac{fb}{d_i}$ where $b$ is the baseline distance between the cameras and $f$ is the focal length. The reconstructed image further undergoes inverse image warping process to retain the color information of the corresponding pixel, discussed later in this paper. 

\begin{figure}[!t]
\centering
\includegraphics[height=5.5cm]{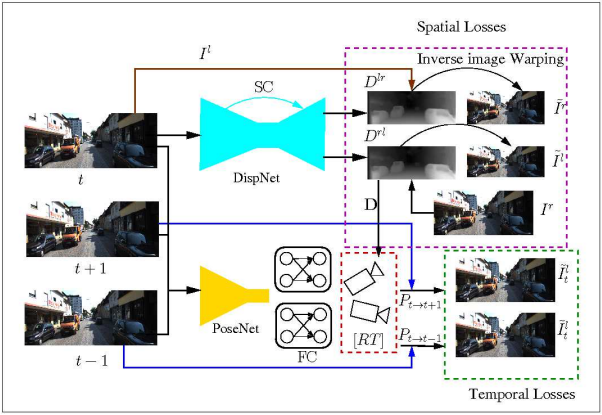} 
\caption{\scriptsize UnDEMoN training procedure. DispNet takes left camera images as input to calculate left-right,right-left disparities $(I^{lr},I^{rl})$. Image warping is applied with predicted disparities to reconstruct left and right images $(\tilde I^l,\tilde I^r)$ with input right,left images $(I^r,I^l)$ respectively. PoseNet takes a snippet of temporally aligned left camera images and reconstructs the target $\tilde{I}_t$ image by predicted 6-DoF pose from source views by applying bilinear interpolation image warping. }
\label{fig:train_meth}
\end{figure}

\subsection{Pose Estimation from Monocular Sequence} \label{pose_estimation}

Given $n$ temporally aligned image sequence  $\{I_1,I_2,..I_t..I_n\}$, our goal is to learn a function $\mathbf{P}$ that can predict frame to frame  6-DoF transformations of the camera. Most of the existing supervised approaches solve this problem by fitting a regression model to the ground-truth camera pose which is often difficult to get. To solve this problem in an unsupervised fashion, we formulate the pose estimation as a temporal image reconstruction problem as described in \cite{dtam}. A temporally aligned image, say, $I_{t+1}$ can be reconstructed (say, $\tilde{I}_{t+1}$ from a given image (say, $I_t$) if the camera transformation from $I_t$ to $I_{t+1}$ and per pixel depth $\textbf{D} = \{\textbf{d}_i, i =1,2,...,N\}$ of current image are known apriori. This can be written as 

\begin{equation}
\label{view_synthesis}
p^{t+1}_i = K P_{t\rightarrow t+1} \textbf{d}_iK^{-1} p^t_i
\end{equation}
where, $K$ is rectified camera calibration matrix. Once the pixel locations are transformed, it would be necessary to compute the pixel color values using differentiable inverse image warping as described next. 

\subsection{Differentiable image warping}
\label{image_warping}
Image warping needs to solve two fundamental problems - one to find the location of the pixel in the reconstructed image and the color values at this location.  The first problem is addressed in the previous two sections. The second problem is address here as explained next. Given a target and source image pair $(I_t,I_s)$ with known transformation ($P$) between the images, the target image can be reconstructed by sampling pixels from the source image through image inverse warping. The warping of image $I_s$ on to the target image $I_t$ is done by interpolating with a sampling kernel in a fully differentiable manner. The image sampler used in this case is obtained from the spatial transformation network (STN)\cite{stn}. If $(u^s_i,v^s_i)$ is the source pixel location in the input feature map and $(u_i^t,v_i^t)$ as the pixel locations of output feature map, then we can use  bilinear sampling kernel of STN that linearly interpolates four neighborhood pixels (top -left, top-right, bottom-left, bottom-right) of source image to approximate the target pixel color using the following equation:
\[
\tilde{I}(u_i^t,v_i^t)=I(u^s_i,v^s_i)=\sum_{i\epsilon\{t,b\},j\epsilon\{r,l\}}w_{ij}I_s(p^s_{ij})
\]
 with $\sum_{ij}{w_{ij}=1}$. The sampling is done independently to each channel in input to preserve the spatial consistency between channels.

\subsection{Network Architecture}
\begin{figure*}[!t]
\centering
\includegraphics[height=6.0cm]{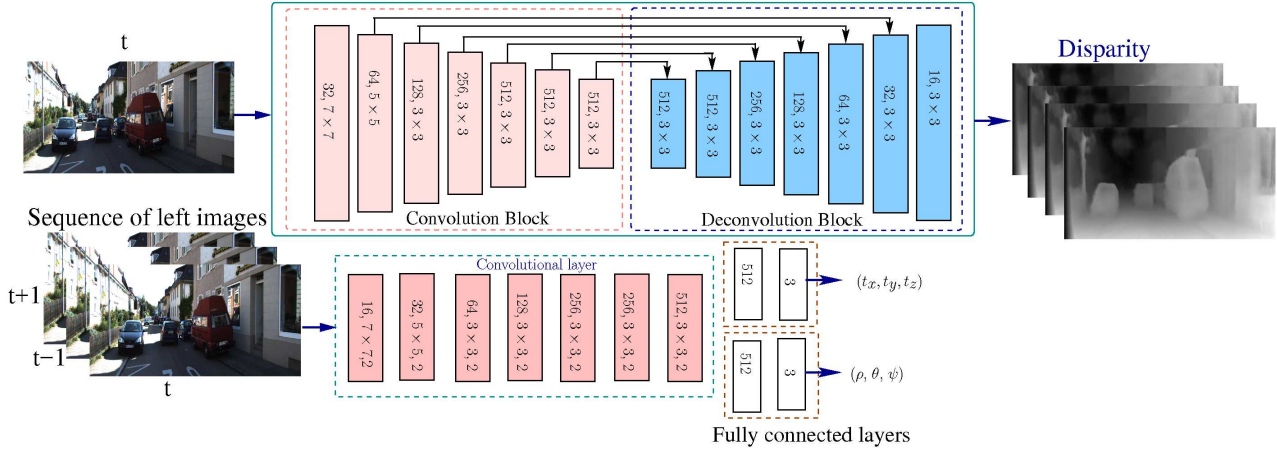} 
\label{landing_fig}\par
\bigskip
\caption{\scriptsize An architectural overview of the proposed UnDEMoN for estimation of depth and camera pose using monocular stereo images. $t^{th}$ frame of the monocular image sequences is used to estimate disparity map and a sequence of temporarily aligned monocular images are given to pose network for predicting image pose in an unsupervised manner. The pose network estimates absolute scale-aware 6-DoF camera pose in terms of rotational and translational parameters: $\rho$, $\theta$, $\psi$ and $t_x$, $t_y$, $t_z$ respectively.}
\label{depth_estimation_nk}
\end{figure*}

Basic architecture of the depth estimation network is taken from the DispNet \cite{disp_net} where the convolutional encoder-decoder is used for multi-scale disparity prediction. We adopt the Rectified Linear Units (ReLU) \cite{relu} as our activation functions to introduce non-linearity for all convolution layers. The prediction layers use sigmoid function $1/(\gamma(sigmoid(D)+\zeta))$ as an activation function. We predict the disparities between $1e-04$ to $d_{max}$, we set $d_{max}$ as $30\%$ of the image width at given output scale ($\gamma=0.3$). The value of $\zeta$ is set to $0.001$. An architectural overview of the proposed technique has been presented in the Fig.\ref{depth_estimation_nk}. Given the left camera image as input, the network can predict left to right disparity $(D^{lr})$ and right to left disparity $(D^{rl})$ in 4 different scales. We have used an encoder followed by fully connected layers in order to estimate camera pose \textbf{P}. The fully connected layers are used for predicting translation $(t_x,t_y,t_z)$ and rotation $(\rho,\theta,\psi)$ independently. The convolutional and fully connected layers in pose network use  ReLU \cite{relu} as activation function except pose prediction layers. The input to the pose network is a sequence of concatenated images and the output of the network gives $6\times(n-1)$ (n is no of input images to pose net) transformations from the target image.

\subsection{Objective Functions} \label{training_losses}

We have trained UnDEMoN using spatial and temporal reconstruction errors. As explained in Section \ref{depth_estimation_sec} and Section \ref{pose_estimation}, the objective function minimizes photometric error between original and reconstructed images. Our objective function is thus framed using two different kinds of losses: appearance loss and disparity smoothness loss. The appearance loss makes sure that the reconstructed image looks like the input image and the disparity smoothness loss ensures that the disparity is always smooth and no discontinuities are present in the predicted disparity. The final loss function is hence given as weighted sum of appearance and smoothness losses $L_{ap}$, $L_{s}$. It is defined as
\begin{equation}
\label{loss}
 L=\lambda_{ap}(L^l_{{ap}}+L^r_{{ap}})+\lambda_{ds}(L^l_{{ds}}+L^r_{{ds}})
\end{equation}
where $L_{ap}$ is the appearance difference loss and $L_s$ is the smoothness loss.

Since we have access to both left and right images, we can calculate all losses using both the images during the time of training. We incorporate a differential variant of absolute norm named Charbonnier penalty \cite{charbonnier_loss} to the objective function. The resultant objective function is a convex function and it is robust to outliers and noises. A generalized Charbonnier penalty is formulated as $\tilde{\rho}=(x^2+\epsilon ^2)^a$.
This penalty function becomes non-convex when $a<0.5$. It has been experimentally found that with a slightly non-convex penalty for $a=0.45$, the performance of the network is better than when $a=0.5$. However, on further change in $a$ values no significant improvements have been observed. The value of $\epsilon$ is set to $0.001$ in all the loss functions.

\subsubsection{Appearance Consistency Loss}

Appearance consistency loss measures the error between the reconstructed and original images. The reconstructed images includes both spatial and temporal reconstructed images. To measure appearance consistency, we tried to measure not only the difference between reconstructed and original images but also structural similarity \cite{ssim}. The structural similarity index is a method of measuring the similarity between images. The similarity index score can be viewed as quality comparison between two images. The index values can be between 1 and 0 where 1 is perfect match 0 is imperfect match. 
The appearance loss can be defined as in eq \ref{apperance_loss}.
Where the $\alpha$ is a weighing factor and is set to 0.85.
\begin{equation}
 \label{apperance_loss}
 L_{ap}=\frac{1}{N}\sum_{ij}\frac{\alpha}{2}\tilde{\rho}(1-SSIM(I_{ij},\tilde{I}_{ij})+(1-\alpha)\tilde{\rho}(I_{ij}-\tilde{I}_{ij})
\end{equation}
The total appearance loss $L_{ap}$ is sum of all appearance losses computed by spatial and temporal reconstruction of images.
\begin{equation}
L^l_{ap}=\lambda_{d}L^d_{ap}(I^l,\tilde{I^l})+\lambda_{p}\sum^{n-1}L^p_{ap}(I^l_t,\tilde{I^l_t})
\end{equation}
Where, $L^d_{ap}$ is the appearance loss between original and reconstructed left image using predicted left-right disparity. $L^p_{ap}$ is the appearance loss between left target image and reconstructed target image from its temporally forward and backward images using equation \ref{view_synthesis}. The constants $\lambda_d, \lambda_p$ are the weights given to the losses. 

\subsubsection{Disparity Smoothness Loss}

We have introduced disparity smoothness loss function to the objective function in order to generate locally smooth disparity map \cite{grad_smooth_loss}. Discontinuity is there in those regions of an image where strong image gradients are present. To avoid such discontinuities, we introduce per pixel exponential weighing function to the disparity gradients $\partial{d_{ij}}$.  
The gradients are calculated by using horizontal and vertical Sobel operators. The gradient smoothness cost can be defined as
\begin{equation}
 \label{gradient_loss}
 L^l_{ds}=\frac{1}{N}\sum_{ij}\tilde\rho({\partial_x d_{ij}e^{-|\partial_xI^l_{ij}|}})+
			    \tilde\rho({\partial_yd_{ij}e^{-|\partial_yI^l_{ij}|}})
\end{equation}

The $\lambda_{ap}$ is set to 1 and $\lambda_{ds}$ is kept to $0.1/s$ where
$s$ is the current image scale value. Here the left camera losses are shown, similarly the same losses are calculated for the right images also. The total sum of left and right losses \ref{loss} is used as the objective function for UnDEMoN.



\section{Experimental Results and Discussions} \label{sec:expt}
In this section we demonstrate and compare the performances of the proposed network for single-view depth estimation and multi-view pose estimation using monocular images. The evaluation and benchmarking is done using KITTI dataset \cite{kitti}. 

\subsection{Dataset and Evaluation Metrics for Depth and Pose Estimation}

The official KITTI dataset \cite{kitti} includes a total of 42,382 rectified stereo image pairs each having a resolution of $1242 \times 375$ taken from 61 scenes in its raw form. The \textbf{KITTI split} divides the total dataset into three sets. The test set includes 200 images taken from 28 scenes, each one having high quality disparity ground truth. The remaining 30095 images from 33 scenes, that do not have any ground truth information, are divided into a training set comprising of 27116 images and a validation set with rest of the 2979 images. 

Similarly, the \textbf{Eigen Split} \cite{eigen} divides all the images taken from 61 scenes into three sets. The test set consists of 697 images taken from 29 scenes. The ground truth for these images is calculated by re-projecting velodyne laser data onto the color image. The remaining 23178 images taken from the other 32 scenes are used for creating the training and the validation set. The training set includes 21055 images and the validation set includes 2123 images.

Apart from the above two splits, there is a third split in this dataset, often called as \textbf{Odometry Split}. It consists of 22 stereo image sequences, saved in loss less png format. Among which, a set of 11 sequences (00 - 10) are provided with ground truth trajectories for training and the remaining images with 11 different sequences ( 11- 21) are given without ground truth for evaluation purposes.

For evaluation, we have used standard scale-invariant metrics such as absolute relative error (Abs Rel), square relative error (Sq Rel), root means error (RMSE) and threshold $\delta$ for depth estimation as defined in prior work \cite{eigen}. Similarly for pose estimation, we are using absolute trajectory error (ATE) \cite{orb_slam} as a performance measure for comparison. 

\subsection{Training details}
We have implemented UnDEMoN using publicly available TensorFlow \cite{tensorflow} framework. A detailed description of the training and testing parameters is given in the table \ref{tab:setup}. 

\begin{table*}[!h]
\centering
\caption{\scriptsize Training and testing setup for UnDEMoN. Training is done using GPU-machine: Quadro-K6000 and testing is done on i-5 laptop. The parameters setting and the training-testing details are given here.}
\label{tab:setup}

\begin{tabular}{S|S|S|S|S|S|S|S} \toprule
    {Training} & {\# parameters} &{Training} &{\#  Training} & {\# iteration} & {Test }& {Test} & {Test image} \\ 
    
         {Setup}& & {time}& {images} &  & {setup}& {time} & {size} \\     \midrule
    {Quadro}  & 35  & 35 &{30}& {$0.2$ }& {Intel} & 70 & {$256 \times 512$}  \\ 
  {K-6000}  & {million}  & {hrs.} &{thousands}& {million}& {i-5} & {ms} &   \\ \bottomrule
  \end{tabular}

\end{table*}
 The proposed UnDEMoN comprises of 35 million parameters that get updated during training. The training image set is of size 30 thousand and it takes around 35 hrs to completely train the model. Adam optimizer \cite{kingma2014adam}, which has been proven to be the best optimization algorithm for deep neural networks, is applied to our network. The training parameters for Adam optimizer are set as $\beta_{1}=0.9$ and $\beta_{2}=0.99$. Initial learning rate is set to 0.001 and it gets reduced by half after completing $(3/5)^{th}$ of total iteration and after $(4/5)^{th}$ of the total iteration, the parameter further gets reduced by half. Initial parameters of the Charbonnier penalty function are set as $\gamma=0.45$ and  $\epsilon=0.001$.

\subsection{Data Augmentation}
To make the estimation model more generic, we apply different kinds of data augmentation techniques with random probability of 0.5. These augmentation techniques include: image flipping, color augmentation and, rotational augmentation. The color augmentation involves applying random brightness sampled in the range of [0.5,2.0], applying random gamma in the range of [0.8,1.2] and randomly shifting colors in the range of [0.8, 1.2]. The application of rotational augmentation degraded the network performance and hence was removed while reporting the final performance in this paper. The details of performance comparison of our proposed method with other state-of-the-art methods is described next in this section below. 
  
  \subsection{Depth Evaluation Results on KITTI and Eigen Splits}

 The performance of our network has been evaluated on KITTI dataset using the Eigen and KITTI splits and, compared with the existing state-of-the-art methods. The performance results on the test sets for both Eigen split and KITTI split are presented in the Table \ref{tab:eigen_split_result} and the Table \ref{tab:kitti_split_result} respectively. The values for existing methods have been taken from their respective papers. As one can observe, our proposed network outperforms all other methods in this category, thereby creating a new benchmark for depth estimation. All the results are computed according to the crops defined in \cite{garg} and \cite{eigen}.
 
\begin{table*}

\caption{Performance Comparison results when validated using KITTI split. The validation set consists of 200 images, taken from 28 different scenes. Each image in the validation set is associated with the disparity ground truth. The presented results of the state-of-the-art technique \cite{mono_depth} is achieved after implementing their code for the same validation set using our hardware setup. The cells in blue color show the accuracy matric (higher the value, better the performance) and the remaining columns give error matrics (lower value gives better performance).}  

\label{tab:kitti_split_result}
\centering
\scriptsize{
\renewcommand{\arraystretch}{1.1}
\begin{tabular}{ |p{1.9cm}|p{1cm}|p{1cm}|p{1cm}|p{1.3cm}|p{1.2cm}|p{1.2cm}|p{1.2cm}|p{1.2cm}|  }
 \hline
  Method & Abs Rel & Sq Rel & RMSE & logRMSE&D1-all & \cellcolor[HTML]{7BABF7} $\delta <$1.25
  &\cellcolor[HTML]{7BABF7}$\delta <$1.25$^2$ & \cellcolor[HTML]{7BABF7}$\delta <$1.25$^3$ \\
 \hline
 
 Monodepth\cite{mono_depth} & 0.124& 1.388& 6.125& 0.217& 30.272& 0.841& 0.936 &0.975\\
 
 \textbf{Ours}    &\textbf{0.1141}  & \textbf{1.1554} &   \textbf{5.831} &  \textbf{0.207} &   \textbf{29.604}
		      & \textbf{0.848}  & \textbf{0.941} &\textbf{0.978}\\

 \hline
\end{tabular}}
\end{table*}

\begin{table*}[!t]
\caption{\scriptsize Performance Comparison of UnDEMoN with existing state-of-the-art techniques using eigen split \cite{kitti}. Results for Liu et al. \cite{learning_depth_mono_cnf} are taken from \cite{mono_depth}. The eigen results are recomputed with velodyne laser data. For fair comparison, the Eigen and Garg results are computed according to crop described in \cite{eigen} and \cite{garg}. The first part of the Table show results of the validation set when Garg crop with 80 meter is used. Similarly, second part and third part show the results for Garg crop with 50 meter and Eigen crop with 80 meter respectively. Also, the second column tells if the techniques have used supervised ($s_{vised}$ depth estimation. It is \emph{Yes}, if is using  supervised approach and \emph{No}, if it is unsupervised. The third column gives information about the pose being estimated for the given technique or not.}
\label{tab:eigen_split_result}
\center
\scriptsize{
\renewcommand{\arraystretch}{1.2}
\begin{tabular}{ |p{2.2cm}|p{0.8cm}|p{0.5cm}||p{1.0cm}|p{1.0cm}|p{1.0cm}|p{1.2cm}|p{1.0cm}|p{1.0cm}|p{1.0cm}|  }
 \hline
  \tiny{Method} &\tiny{S-vised}& \tiny{Pose} & \tiny{Abs Rel} & \tiny{Sq Rel} & \tiny{RMSE} & \tiny{RMSE log} & \tiny{\cellcolor[HTML]{7BABF7} $\delta <$1.25}
  &\tiny{\cellcolor[HTML]{7BABF7}$\delta <$1.25$^2$} &\tiny{ \cellcolor[HTML]{7BABF7}$\delta <$1.25$^3$} \\
 \hline
 Train set mean &No &No& 0.361& 4.826& 8.102& 0.377& 0.638& 0.804& 0.894\\

 Eigen et al. & Yes& No& 0.214& 1.605& 6.563& 0.292& 0.673& 0.884 &0.957\\
Coarse \cite{eigen} & & & & & & & & &\\

 Eigen et al.  & Yes &No& 0.203& 1.548& 6.307& 0.282& 0.702& 0.890& 0.958\\
Fine \cite{eigen} & & & & & & & & &\\

Liu et al. \cite{learning_depth_mono_cnf}& Yes& No& 0.201& 1.584& 6.471& 0.273& 0.68& 0.898& 0.967\\
 SfM Learner \cite{sfm_learner}&No& Yes& 0.208& 1.768& 6.856& 0.283& 0.678& 0.885& 0.957\\
Mono depth \cite{mono_depth}& No& No& 0.148&   1.344 &  5.927 &  0.247   &   0.803   &   0.922 &    0.964\\ 
 Undeep VO \cite{undeepvo}&No& Yes & 0.183 & 1.73& 6.57& 0.268 &     - &    - &    -\\
 
 \textbf{Ours}         &No&Yes&\textbf{0.1396}  & \textbf{1.1497} &   \textbf{5.571} &  \textbf{0.238} &   
 
 \textbf{0.810}  & \textbf{ 0.930} &  \textbf{0.968} \\

 
 \hline
 Garg et al \cite{garg} cap 50m& No &No& 0.169& 1.080& 5.104& 0.273& 0.740& 0.904& 0.962\\
 SfM Learner\cite{sfm_learner} cap 50m& No&Yes& 0.201& 1.391& 5.181& 0.264& 0.696& 0.900& 0.966\\
 Mono depth \cite{mono_depth}cap 50m& No& No& 0.1402 &  0.9764 & 4.471  &  0.232 &     0.818  &    0.931  &   0.969\\
\textbf{Ours}    &No&Yes&\textbf{0.1323}  & \textbf{0.8846} &   \textbf{4.290} &  \textbf{0.226} &   \textbf{0.827}
		      & \textbf{ 0.937}  & \textbf{ 0.972} \\
 		      
  \hline
 Mono depth \cite{mono_depth} eigen crop& No& No& 0.1691&   1.7212 &  6.538 &  0.269   &   0.770   &   0.906 &    0.955\\
 \textbf{Ours}         &No&Yes&\textbf{0.1567}  & \textbf{1.4617} &   \textbf{6.136} &  \textbf{0.259} &   \textbf{0.782}
		      & \textbf{    0.916}  & \textbf{ 0.961} \\ 
  \hline
\end{tabular}}

\end{table*}


\subsection{Pose Evaluation Results on Eigen splits}

Since the model is trained on the KITTI Eigen split for depth estimation, the pose evaluation is also carried out on the same split. We have found that four of the odometry split scenes with ground truths are not included in the Eigen training split and hence, can be used for testing the pose estimation performance. The performance of pose estimation is quantified using Absolute Trajectory Error (ATE) \cite{orb_slam} for both rotation and translation. Table \ref{tab:kitti_split_result} shows the performance comparison of the proposed network with another state-of-the-art deep learning method called SfMLearner \cite{sfm_learner} and one conventional pose estimation method VISO \cite{libviso}. It is to be noted that the results directly reported in \cite{sfm_learner} are obtained after post-processing. This is reported in the table under the heading SfMLearner\_PP. For a fair comparison, we re-run the algorithm by removing this post-processing module to obtain results for the column with heading SfMLearner\_noPP. Similarly, we report the results for both monocular and stereo version of VISO algorithm. This Table shows that UNDEMoN outperforms both SfMLearner\_noPP and VISO\_M.  At the same time, its performance is comparable with the methods that either use post-processing for retrieving scale as in SfMLearner\_PP \cite{sfm_learner} or use stereo images for pose estimation as in VISO\_S \cite{libviso}.

As the KITTI training dataset has very few scenes with rotational variation, the rotation component is not learnt properly by the deep network, leading to large accumulated pose error over long distances as shown in Figure \ref{fig:pose_results}. The rotational data augmentation does not solve this problem and needs to be investigated further.

\subsection{Discussion on Performance Evaluation}
Even though we compare the performance of our network with those of UnDeepVO \cite{undeepvo}, which is the latest work in this field,  in Tables \ref{tab:eigen_split_result} and \ref{tab:kitti_split_result}, such a comparison may not be fair as we notice several inconsistencies in the results reported in their paper.  For instance, for depth estimation, the network is trained on the odometry split while the test results are shown for the eigen split dataset which itself includes few scenes from the odometry split.  This appears to be inconsistent with the standard practice of using different sets for training and testing. Similarly, the authors show the performance of pose estimation by presenting only the training results which, though appears good, are again inconsistent with the standard practice of reporting results on a separate test dataset.   In contrast, we are reporting results only for test datasets which are different from the training datasets and still, we outperform their training results for depth and pose estimation. We also plan to publish our source codes post the acceptance of this paper. 

\begin{table*}
\caption{\scriptsize Absolute Trajectory Error (ATE) \cite{orb_slam} for Translation and Rotation on KITTI eigen split dataset averaged over all 3-frame snippets (lower is better). As one can see, UnDEMoN outperforms the monocular versions SfMLearner\_noPP and VISO\_M and is comparable with stereo versions SfMLearner\_PP \cite{sfm_learner} and VISO\_S \cite{libviso}. Here, the terms $t_{ate}$ and $r_{ate}$ stand for translational absolute trajectory error and rotational absolute trajectory error respectively.}
\begin{center}
\scriptsize{
\label{tab:ate}
\begin{tabular}{|c|cc|cc|cc|cc|cc|} \toprule[1.5pt]
 \head{Seq} & \multicolumn{2}{c|}{UnDEMoN} & \multicolumn{2}{c|}{SfMLearner\_noPP \cite{sfm_learner}} & 
    \multicolumn{2}{c|}{SfMLearner\_PP \cite{sfm_learner}} & \multicolumn{2}{c|}{VISO2\_S \cite{libviso}} & \multicolumn{2}{c|}{VISO\_M \cite{libviso}}\\
  & $t_{ate}$ &  $r_{ate}$ & $t_{ate}$ &  $r_{ate}$ &  $t_{ate}$  &   $r_{ate}$ & $t_{ate}$ &   $r_{ate}$
& $t_{ate}$ &   $r_{ate}$   \\ \hline 
  00 &  0.0644 & 0.0013 & 0.7366 & 0.0040 & 0.0479 & 0.0044 & 0.0429 & 0.0006 & 0.1747 & 0.0009\\
  04 &  0.0974 & 0.0008  & 1.5521 & 0.0027 & 0.0913  & 0.0027  & 0.0949 & 0.0010 & 0.2184 & 0.0009\\
    05 &  0.0696  & 0.0009 & 0.7260 & 0.0036 & 0.0392  & 0.0036  & 0.0470 & 0.0004 & 0.3787 & 0.0013\\
     07 &  0.0742 & 0.0011 & 0.5255 & 0.0036 & 0.0345 & 0.0036  & 0.0393 & 0.0004 & 0.4803 & 0.0018\\
  \bottomrule[1.5pt]
\end{tabular}}
\end{center}
\end{table*}
\section{Ablation studies} \label{sec:ablation}
We have performed various ablation studies to justify the choice of our Charbonier penalty based objective function and to validate different parameter choices in our proposed framework. All the results are obtained using the KITTI dataset. An empirical analysis for choosing the Charbonier penalty as 0.45 is presented in the Table \ref{tab:ablation_char}. Experiments are carried out for a set of parameter choices, such as 0.35, 0.4, 0.45, 0.5 and 0.55. The performances of two parameters, 0.4 and 0.45 are observed to be distinguishably superior then the other given parameters. Among these two choices, we have opted to use 0.45 as the Charbonier penaly, as it is found to be performing comparatively better for both depth and pose estimation networks.   
\begin{table*}
\caption{Performance evaluation studies on Eigen Split for different Charbonier penalty parameters. Observations show that
choice of the Charbonier penalty as 0.45 provides best depth evaluation results. }
\label{tab:ablation_char}
\centering
\scriptsize{
\renewcommand{\arraystretch}{1.1}
\begin{tabular}{ |p{1.9cm}|p{1cm}|p{1cm}|p{1cm}|p{1.3cm}|p{1.2cm}|p{1.2cm}|p{1.2cm}|p{1.2cm}|  }
 \hline
  Method & Abs Rel & Sq Rel & RMSE & logRMSE&D1-all & \cellcolor[HTML]{7BABF7} $\delta<$1.25
  &\cellcolor[HTML]{5BABF7}$\delta<$1.25$^2$ & \cellcolor[HTML]{7BABF7}$\delta<$1.25$^3$ \\
 \hline
Chabonier 0.35 & 0.1428 &    1.3553 &      5.777&       0.239&      0.000&      0.815&      0.929&      0.967\\
\rowcolor[HTML]{7BCBE7}
 charbonier 0.4 &    0.1396 &     1.2433 &      5.643 &      0.238 &      0.000 &      0.814 &      0.930 &      0.968\\
\rowcolor[HTML]{7BCBE7}
 charbonier 0.45 &  0.1396 &     1.1497&      5.571&      0.238&      0.000&      0.810&      0.930&      0.968 \\
 
 charbonier 0.5 &  0.1408&     1.3078&      5.758&      0.240&      0.000&      0.815&      0.929&      0.967 \\
charboner 0.55 &  0.1436 &     1.2479 &      5.738 &      0.245 &      0.000 &      0.805&      0.926&      0.965 \\
 \hline
\end{tabular}}

\end{table*}
Next analysis we have performed on the choice of snippets and the different evaluation results for both pose and depth are presented in the Table \ref{tab:ablation_snippet_pose} and Table \ref{tab:kitti_ablation_snippet_depth} respectively. Totally, we have used 3 different types of snippets: 2-frame, 3-frame (presented previously in this paper) and 5-frame. In the work \cite{madhu-iros}
, we have used only 3-frame snippet for the experiments. However, extensive analyses with the two more snippets show that, the pose estimation results are better on using 2-frame snippet than of that the 3-frame snippet, whereas the depth estimation results are superior on using 3-frame snippet. 5-frame snippet gives worse estimation performances for both the networks. Such conflicting observations could be because of the following reasons: As we are are moving from 2-frame snippet to 3-frame snippet the depth network could able to capture more temporal information, however, on further increasing the number of temporal informations (frames) both in forward and backward directions, more ambiguities gets added into the network. The ambiguities are mostly background noises, i.e, as an object (car in this case) is moving forward, other background objects may also be moving in various directions ( need not be static). These actually, misleads the estimation network.

\begin{table}
  \captionsetup{width=0.90\linewidth}
  
  \caption{\scriptsize Absolute Trajectory Error (ATE) \cite{orb_slam} for Translation and Rotation on KITTI eigen split dataset averaged over two different snippets: 2-frame snippets  and 5-frame. It can be observed, that the pose estimation result using 2-frame snippet is better than the 5-frame snippet and the 3-frame snippet presented in the Table \ref{tab:ate}.
   Here, the terms $t_{ate}$ and $r_{ate}$ stand for translational absolute trajectory error and rotational absolute trajectory error respectively.}
   \label{tab:ablation_snippet_pose}
   \vspace{5 pt}
   \centering
   \footnotesize{
\begin{tabular}{ccccc}
\toprule 
Seq & $t_{ate}$ &  $r_{ate}$  & $t_{ate}$ & $r_{ate}$ \\
 & (2- snippet) &  (2- snippet)  & (5- snippet) & (5- snippet) \\
\midrule
{00} & 0.0481 & 0.0010 & 0.1356 &  0.0026 \\
\midrule
{04} & 0.1269 & 0.0004 &0.2544  &  0.0014  \\
  

\midrule
 {05} & 0.0486 & 0.0006 & 0.1266 & 0.0018  \\
\midrule

 {07} & 0.0562 &0.0007 &   0.1468 & 0.0021  \\
  
\bottomrule
\end{tabular}}
\end{table}

\begin{table*}
\caption{Depth estimation results on KITTI split when 2-frame snippet and 5-frame snippet are being used. The validation set consists of 200 images, taken from 28 different scenes. Each image in the validation set is associated with the disparity ground truth. The cells in blue color show the accuracy matric (higher the value, better the performance) and the remaining columns give error matrics (lower value gives better performance).}  

\label{tab:kitti_ablation_snippet_depth}
\centering
\scriptsize{
\renewcommand{\arraystretch}{1.1}
\begin{tabular}{ |p{1.9cm}|p{1cm}|p{1cm}|p{1cm}|p{1.3cm}|p{1.2cm}|p{1.2cm}|p{1.2cm}|p{1.2cm}|  }
 \hline
  Method & Abs Rel & Sq Rel & RMSE & logRMSE&D1-all & \cellcolor[HTML]{7BABF7} $\delta <$1.25
  &\cellcolor[HTML]{7BABF7}$\delta <$1.25$^2$ & \cellcolor[HTML]{7BABF7}$\delta <$1.25$^3$ \\
 \hline
  
 2-frame snippet & 0.1454& 1.2410& 5.789& 0.247& 0.000& 0.800& 0.922 &0.964\\
 
 5-frame snippet & 0.1437& 1.4263& 5.785& 0.239& 0.000& 0.816& 0.929 &0.967\\

 \hline
\end{tabular}}
\end{table*}


\section{Conclusion}\label{sec:conc}

This paper looks into the problem of estimating depth and camera pose from monocular stereo images. The problem is solved by using a  deep network framework comprising of a depth estimation network and a pose estimation network. The network is trained in an end-to-end unsupervised fashion by using a novel objective function based on Charbonnier penalty applied to both spatial and bi-directional temporal reconstruction losses. The proposed deep network named as "UnDEMoN" incorporates best features of the existing architectures and methods to provide improved depth and pose estimation performance that beats the current state-of-the-art. However, the current work does not address the issues arising out of moving objects in a scene which may affect the estimation performance adversely. Moreover, the pose estimation performance of our proposed network could further be improved by training on a dataset that has more rotational variations unlike KITTI dataset that has very few such scenes. It would also be interesting to see if the deep network could be trained using monocular images along with noisy IMU data to estimate depth and pose. This will obviate the need for stereo image pairs currently needed for our network. Many of these concerns will guide our future direction of this research. 

\bibliographystyle{abbrv}

\end{document}